# African Gender Classification Using Clothing Identification Via Deep Learning

Samuel Ozechi

Lagos, Nigeria

## Abstract

Human attribute identification and classification have become essential aspects of computer vision, contributing to the development of innovative systems in recent years. Traditionally, gender classification systems heavily rely on facial recognition techniques, which, while effective, often suffer limitations under non-ideal conditions such as blurred images, side views, or partial obstructions. This study explores an alternative approach to gender classification by focusing on clothing identification, specifically using traditional African attires rich in cultural and gender-specific features. African traditional clothing is uniquely tied to ethnic identity, social status, and gender, offering a more contextually relevant and reliable basis for gender classification. This work utilizes the AFRIFASHION1600 dataset, a contemporary collection of 1,600 images of African traditional attire, engineered to include two gender classes—male and female. A deep learning model was trained using transfer learning techniques with a modified VGG16 architecture. Data augmentation was applied to mitigate the challenges posed by the relatively small dataset size and prevent overfitting. The model achieved an accuracy of 87% on the test set, demonstrating strong predictive capability despite inherent dataset imbalances favoring female samples. This study highlights the potential of clothing identification as a complementary technique to facial recognition for gender classification in African contexts. The results underscore the need for further work, including the inclusion of more diverse and balanced datasets, to build a more robust and innovative gender classification system.

## 1. Introduction

Facial recognition systems have mostly been employed for gender classification in computer vision over the years. The development of advance tools and concepts in deep learning and computer vision, especially with developing convolutional neural networks to construct maps of facial features or utilizing transfer learning of models pretrained on huge volume of data, has overtime, further enhanced the credibility and resultant reliability of using facial recognition techniques for many forms of attribute identification in computer vision.

However, the use of facial recognition for attributes identification and classification are not without pitfalls as most facial recognition techniques require clear images and high-quality videos of clearly defined facial features to be effective [5]. In real life situations, facial images, especially when obtained under non ideal situations, are often distorted, blurred or concentrated on positions that are not appropriate for proper feature detection by facial recognition techniques. For example, common facial recognition techniques are often unable to detect facial features on images where the face is inside view, taken from distance, partially covered or even blurred.

While new innovative techniques, like Google's Mediapipe facial recognition system are addressing some of these limitations, these are still valid problems today for most facial recognition techniques such as the OpenCV Facial Cascades.

In contrast to facial recognition, clothing identification offer less limitations in gender classification. Though not entirely new, as clothing identification have been employed by humans for gender classification over time immemorial [10], it offers an improved, simple and larger dimensions for gender identification. It is simpler to assume a human wearing a skirt is female and another wearing a Tuxedo is male.

Even though there are unisex clothing with less defined gender boundaries like shirts and trousers which pose classification problems especially in European fashion trends, gender classification by clothing identification is even narrower in the African fashion system as there are only but a few unisex traditional attires or even none for some African cultures. The Yoruba ethnic society for instance, have distinct cultural attires for both the male and female genders, just like most African cultures, making it possible to rather easily identify genders in traditional attires, even without facial detection, from images taken at relative distances and partially obstructed.

Another reason why clothing identification for gender classification seems quite appropriate for Africans is that most African cultures significantly recognize only the male and female gender groups, thereby limiting the misclassification that may arise from non-binary gender classes.

Therefore, while gender classification is made complex by the ever-increasing gender classes, the more generic male and female gender class system of African cultures simplifies the gender distinction which in result is a boost for classification with deep learning techniques as it eliminates the possibility of multi-label classification complexity.

Rather than totally replace facial recognition systems, clothing identification is best utilized for complementing facial recognition techniques [5] and further enhancing the credibility of human attribute identification using computer vision.

This project aims to build a model using deep learning, to classify the gender of Africans based on their common traditional attires.

## 2. The Dataset

There are quite a few fashion datasets that have influenced the advances in clothing identification [7]. The most popular being the Fashion MNIST (Modified National Institute of Standards and Technology) dataset by Xan Hiao et al [4] of 70000 images, the Deep fashion and fashion landmark datasets of 800000 and 120000 images respectively by Liu et al [8, 9] and the Modanet dataset of more than 55000 images by Zheng et al [6].

Due to the Afrocentricity of this project, none of the above listed datasets is appropriate for this project, as it requires a dataset that represents African traditional attires, hence the AFRIFASHION1600 dataset was chosen.

The AFRIFASHION1600 dataset, introduced by Wuraola et al [7] in 2021, is currently the only notable contemporary Afrocentric fashion dataset. It contains 1600 (180 x 180 dimensions) RGB sample images that are categorized into 8 classes of fashion styles.

The dataset was further engineered to include 2 gender classes (female and male) that are specific and relevant to this project.

The final form of the dataset used for this project contains a directory of 1600 images of African attires and a comma separated file (CSV) that includes the image IDs, fashion style and gender class columns.

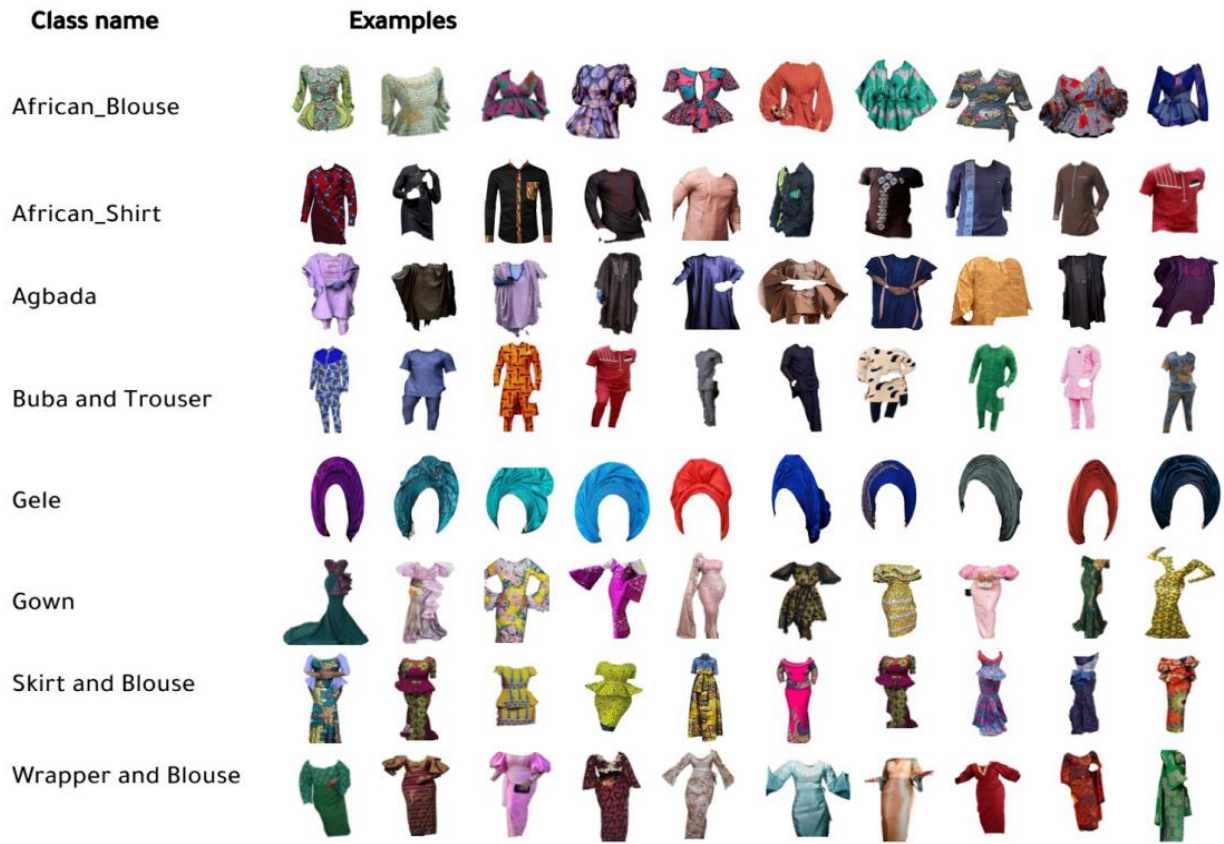

Adapted from [7]

*Figure 1. Class names and sample images for the AFRIFASHION1600 dataset.*

|  | image_id | clothing | gender |
|---|---|---|---|
| 0 | African_Blouse_1.png | African Blouse | Female |
| 1 | African_Blouse_10.png | African Blouse | Female |
| 2 | African_Blouse_100.png | African Blouse | Female |
| 3 | African_Blouse_101.png | African Blouse | Female |
| 4 | African_Blouse_102.png | African Blouse | Female |
| ... | ... | ... | ... |
| 1595 | Wrapper and Blouse_95.png | Wrapper and Blouse | Female |
| 1596 | Wrapper and Blouse_96.png | Wrapper and Blouse | Female |
| 1597 | Wrapper and Blouse_97.png | Wrapper and Blouse | Female |
| 1598 | Wrapper and Blouse_98.png | Wrapper and Blouse | Female |
| 1599 | Wrapper and Blouse_99.png | Wrapper and Blouse | Female |

1600 rows × 3 columns

*Figure 2. Preview of image_Ids, fashion labels and gender classes in a dataframe.*

Exploratory analysis shows that 62.5% of the dataset belonged to the female class while 37.5% belonged to the male class which results in a data imbalance. It is thus expected for the trained classifier to make a biased learning model that could have a poorer accuracy on the male class compared to the female class.

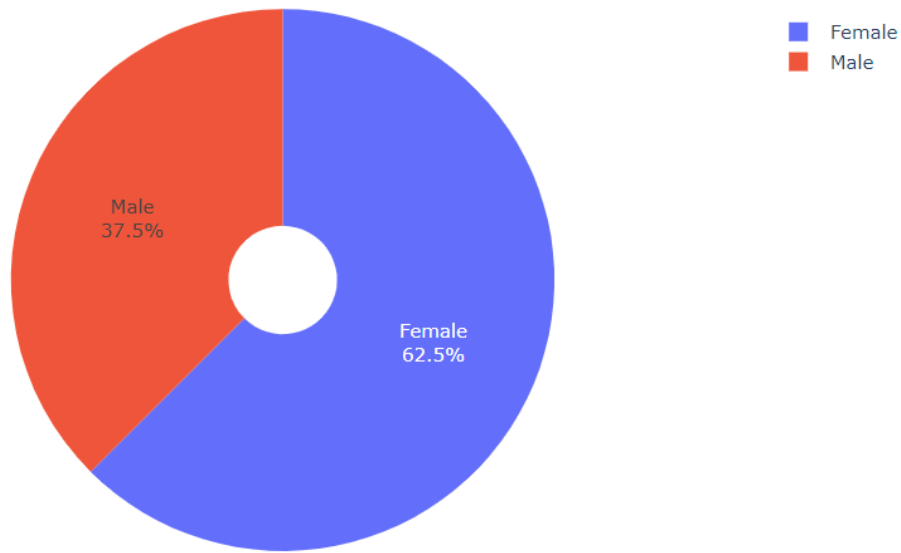

*Figure 3. Pie Plot of the gender distribution of the dataset.*

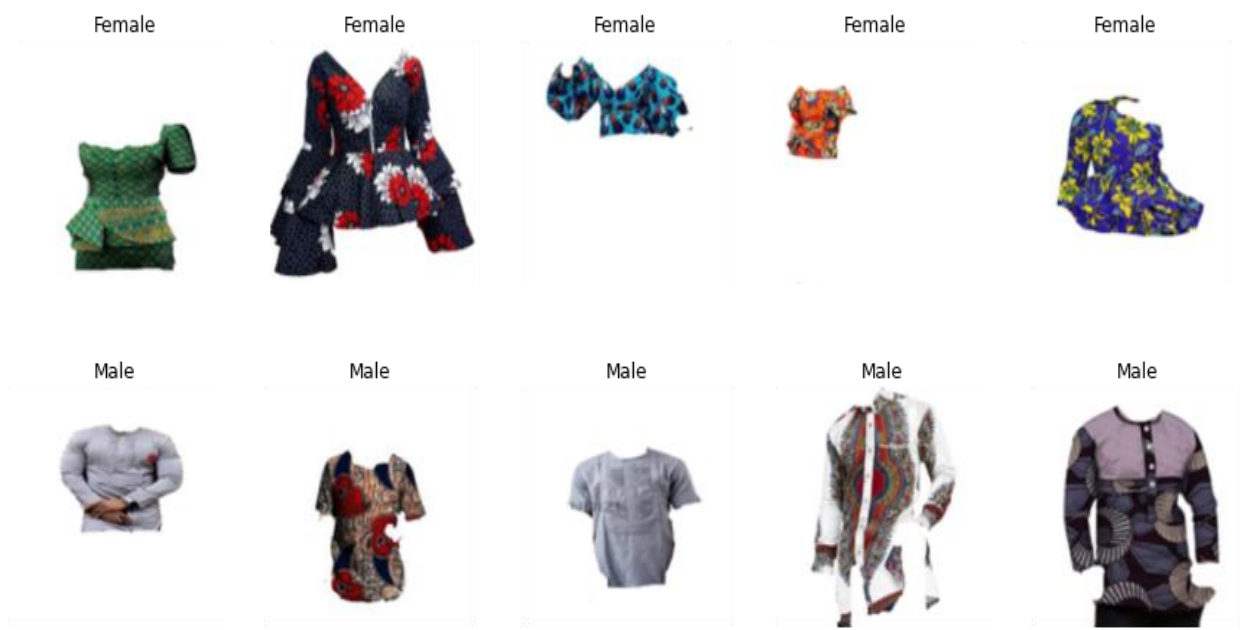

*Figure 4. Sample images from the dataset and their gender labels.*

## 3. Methods

Deep learning image classification techniques were utilized for this project.

80% of the entire data were allocated for training while 10% each were set aside for validating and testing the model.

### 3.1. Data Preprocessing

Due to the relatively low number of samples, the image data was preprocessed by applying data augmentation to the training set to enhance the model accuracy and limit overfitting.

Data augmentation is an image preprocessing technique utilized in computer vision to augment image data of fewer samples. This technique takes the approach of generating more training data from existing training samples, by augmenting the samples via a number of random transformations that yield believable-looking images [3]. The goal of data augmentation is to expose the model to more aspects of the data and generalize better.

This was implemented using the *Image Data Generator* instance of the Keras library.

The *Image Data Generator* instance was used to load the images as array values, apply data augmentation and store the data in generator objects (train, validation and test generators) using a batch size of 128.

### 3.2. Model Architecture and Classification

Transfer learning technique is utilized for the model building and classification in this project. This involves the use of saved Convnet models with weights that have have been pretrained on large datasets. The spatial hierarchy of features learned by pretrained networks have proved useful in other computer vision problems, even with different class targets [3].

The saved weights of the VGG16 model, pretrained on the Imagenet dataset were specifically used in this project. The Visual Geometry Group 16 (VGG16) is a convolutional neural network proposed by A. Zisserman [1]. Convolutional neural networks are deep learning models that are often used for computer vision problems. They are networks of deeply connected neural layers that perform spatial filtering, convolution, back-propagation and gradient descent operations on image inputs for image classification.

The input to the first layer (Conv1) of the original VGG16 model is of a fixed size 224 x 224 Red, Green, Blue (RGB) image. The image is passed through a stack of 24 layers, with series of 3x3 and 1x1 shaped filters. Spatial pooling is carried out by 5 Maxpooling layers (of size 2x2 and stride of 2) which follows only some of the Convolution layers.

The final two layers are a 1000 channels Convolution layer (for each class of the Imagenet dataset) and a Softmax layer for multi-class classification [1].

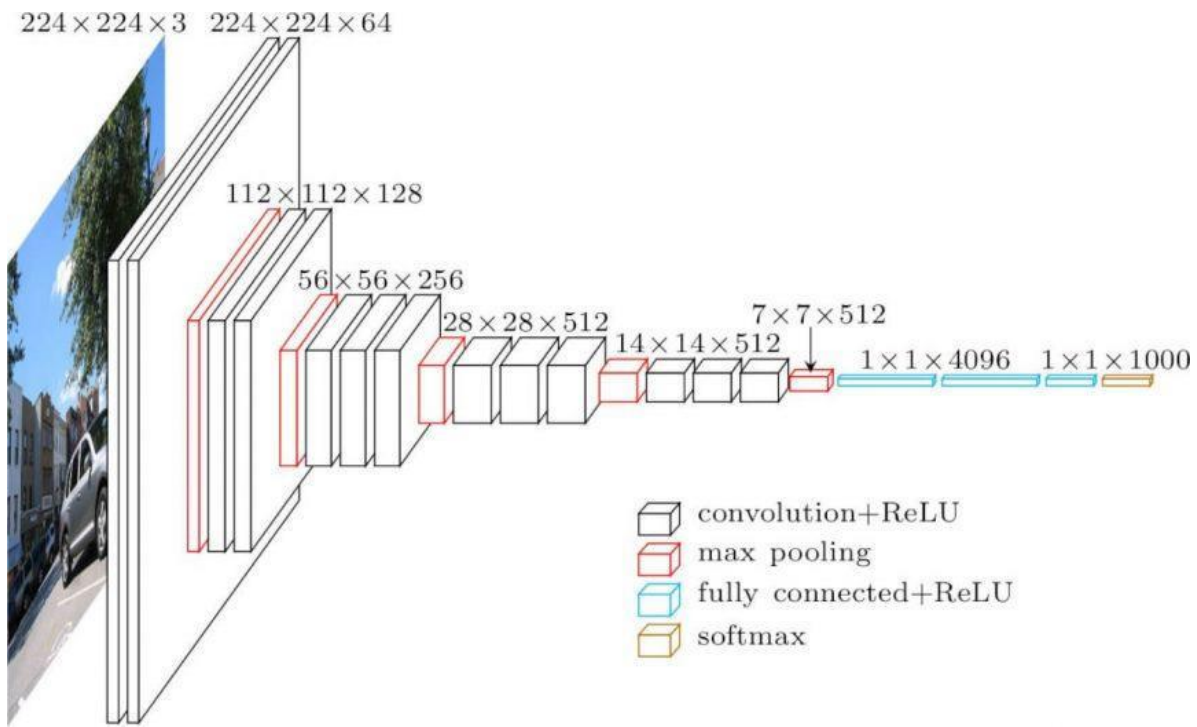

*Figure 5. VGG16 model architecture. Adapted from [1].*

The top layers of the model were not included due to the difference in the number of classification targets and size of the input images. The last four (4) layers of the model were also unfreezed to allow the model learn features specific to this image classification problem while utilizing the features learned from the Imagenet dataset.

Newly defined flatten, dropout (for regularization) and output layers (with Sigmoid activation function) were added to the modified VGG16 model.

Sigmoid activation functions are commonly used for binary classification to map the probability of outputs between 0 and 1, with positive outcomes for values greater than 0.5 and negative for those below.

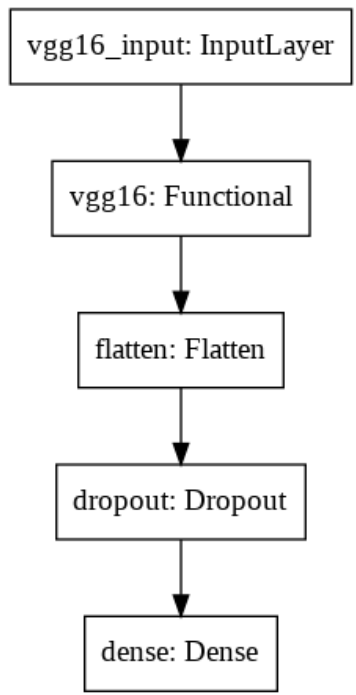

*Figure 6. Architecture of the model.*

### 3.3. Model Training

The training sets, now stored in data generators of 128 batch sizes were used to train the model for 50 epochs while validating the model using the validation set with training and validation steps of 13 iterations per epoch. The batch size is the number of samples that are propagated through the network for each forward pass and weight adjustment. The batch size helps to implement the mini batch gradient descent which is used for this project.

The training and validation steps (result of dividing the number of samples by the batch size) defines the number of iterations it takes to propagate all the samples through the back propagation algorithm, given it propagates a batch of the samples in each iteration.

The epochs describe the number of times the algorithm process the entire dataset. An epoch is completed when the number of steps required to propagate the entire dataset at batches is complete.

Callbacks were defined to utilize monitoring the validation loss for saving the best weights, reducing the learning rate at intervals and stopping the training if there weren't any learning after few epochs.

### 3.4. Model Evaluation

After 50 epochs of training, the model achieved a 90% accuracy on the train set with a minimum loss of 0.18 and an 89% accuracy on the validation set with a minimum loss of 0.25.

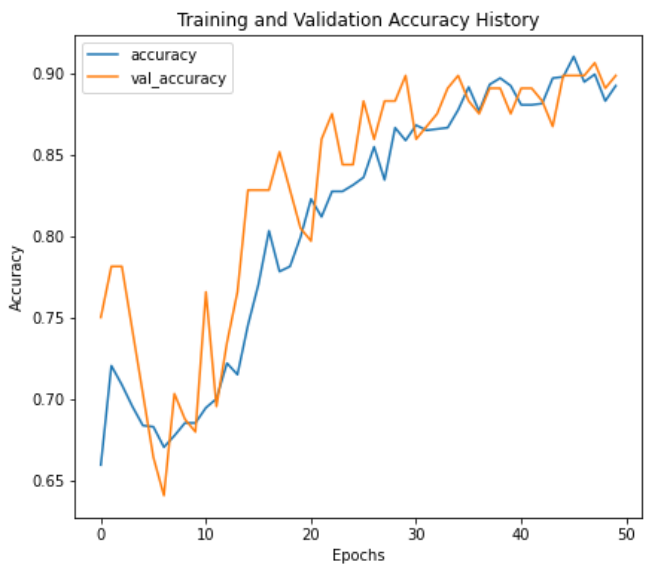
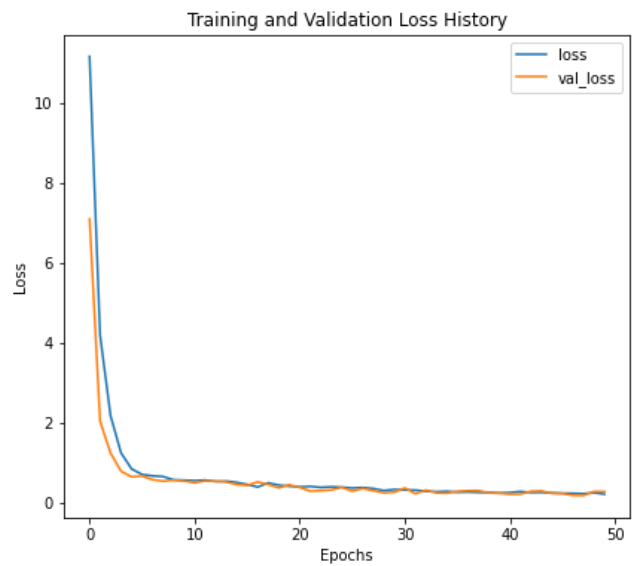

*Figure 7. Performance of the model on the train and validation sets.*

The model also achieved an 87% accuracy on the test set. A confusion matrix evaluation showed that while the model is biased towards the female class, it still showcased good predictiveness for both classes.

The confusion matrix is a useful visualization of performance measurement for classification algorithms.

It shows the model's performance in terms of true and false positive and negatives in the test set evaluation.

The true positive is the number of correctly predicted samples for the positive class while the false positive is the number of wrongly predicted samples for the same positive class (female).

The false negative is the number of wrongly predicted samples for the negative class (the not female class, which is the male class) while the true negative is the number of rightly predicted samples for the same negative class.

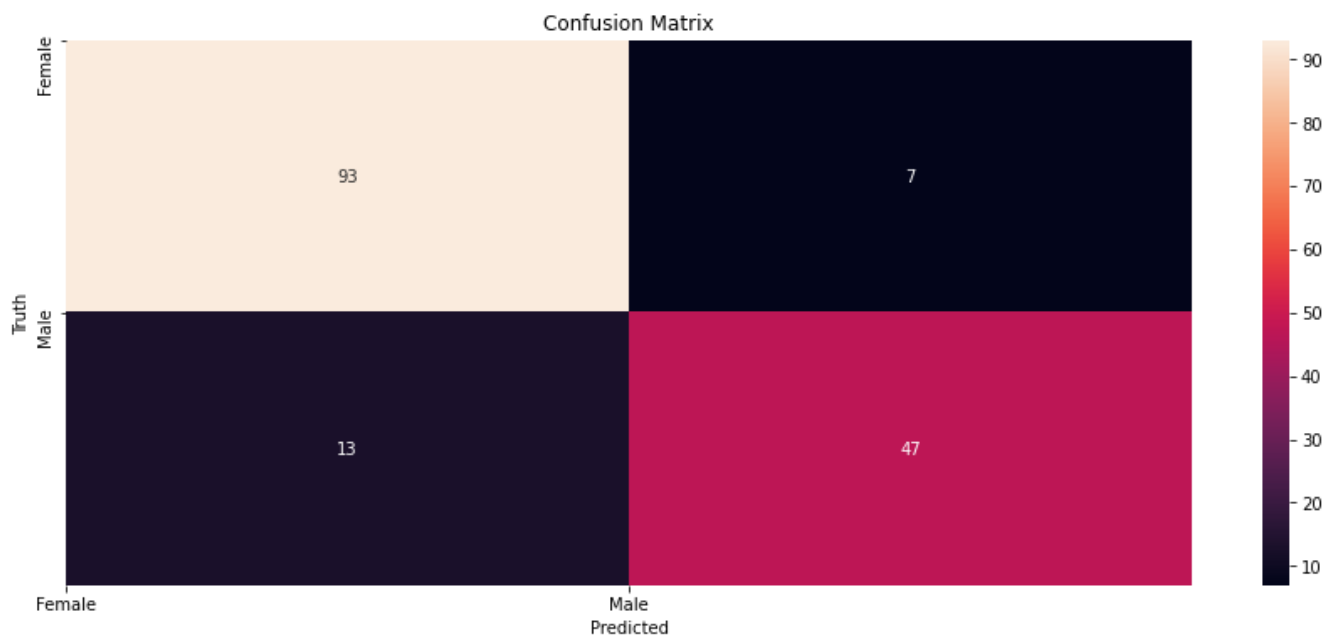

*Figure 8. Confusion matrix of test set evaluation.*

Only 7 (false positive) out of the 100 female test samples were misclassified (93% precision) while 13 (false negative) of the 60 male test samples were misclassified (78% precision). This disparity in precision is because of data imbalance.

A classification report is also useful in measuring the quality of the prediction from a classification algorithm.

The precision defines the ratio of the true positives or correct prediction to the number of samples in the class.

The recall defines the ratio of the true positive instances to the sum of the true positives and false negative. It describes the ability of the classifier to correctly predict positive instances.

The F1 score describes the ratio of correct positive instances. It is the harmonic mean of the precision and recall.

```
Classification Report
              Recall    Precision    f1-score    support

       0       0.88        0.93        0.90         100
       1       0.87        0.78        0.82          60

accuracy                               0.88         160
macro avg      0.87        0.86        0.86         160
weighted avg   0.87        0.88        0.87         160
```

*Figure 9. Classification report for the test set evaluation.*

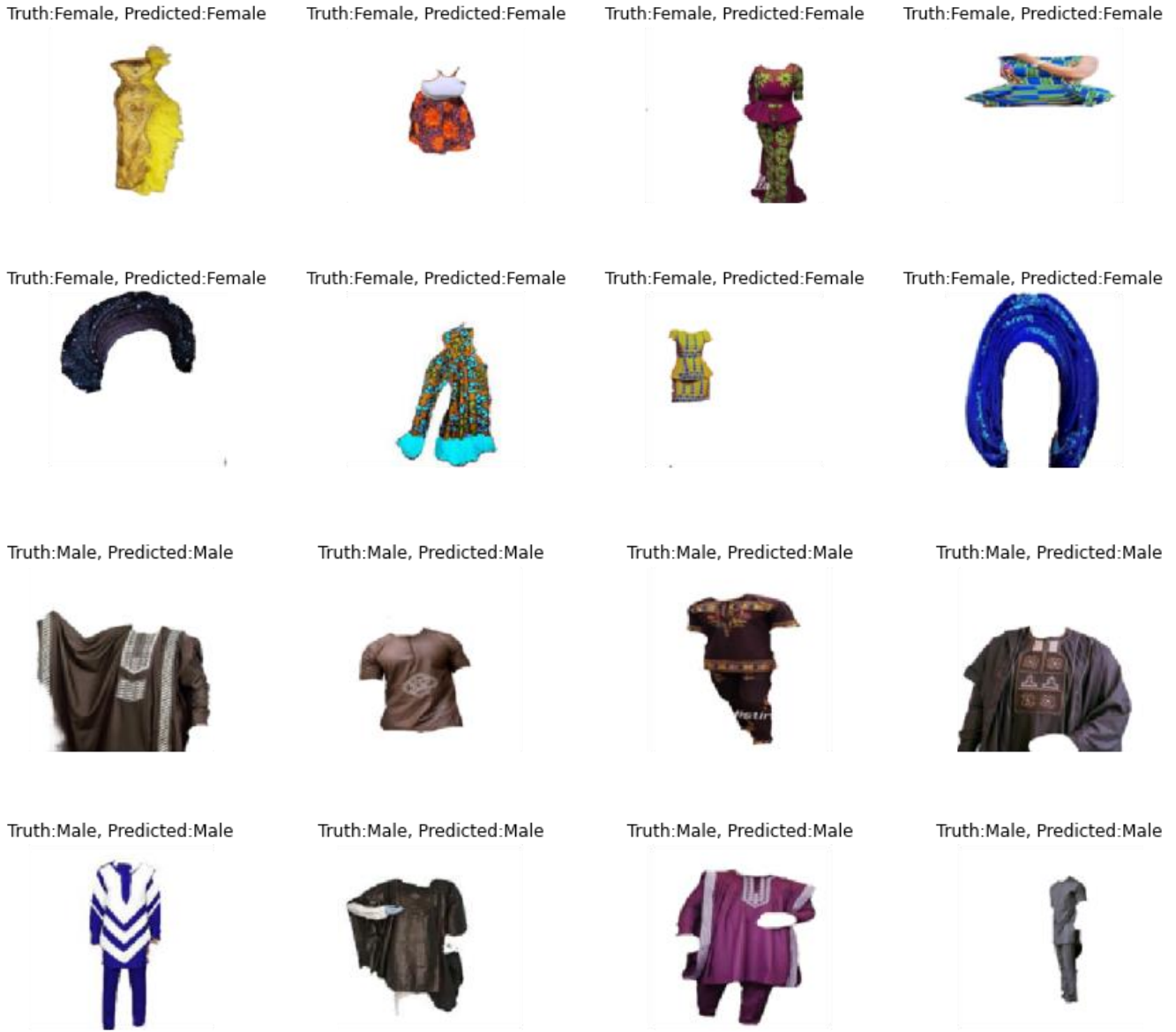

*Figure 10. Comparing the truth and predicted labels for sample images.*

## 4. Discussions

The model's evaluation metric values are useful in measuring the performance and reliability of the model. The average scores of 89% and 87% on the validation and test sets respectively proves a high predictive ability of the model in identifying the clothing for both genders.

The model's precision values for both classes show that the model performs better in predicting the correct class of female fashion samples than for male samples (with precision scores of 93% to 78%). This is because of imbalance in the dataset (62.5% to 37.5%).

The relatively low loss values for both validation and test sets and high recall values suggest that model is less prone to misclassify the fashion samples, especially for the dominant female class whose performance contributes to the desired loss value than the male class.

The F1 scores of 90% for the female class and 82% for the male class are good enough to suggest a decent credibility and robustness of the model as it considers a balance between the precision and recall metrics.

These metric values also mean that the model is a good fit and appropriate to train on the data as it only slightly overfits, helped by the dropout regularization technique utilized in building the model.

Also, despite the training set having few samples (1280 samples) to properly train an image classifier using deep learning, the relatively high metrics scores achieved could be attributed to the learned weights of the pretrained model, which trained on the Imagenet dataset that also contained fashion samples and to the clearly defined, high quality sample images used in training the model.

While the evaluation sets (validation and test sets) relatively have fewer values to properly ascertain the credibility and robustness of the model, it could be suggested that the model performs fairly well in rightly classifying the gender labels for traditional African attires at least.

There is no doubt that the model would greatly benefit from training from much more samples, especially for the less represented male class.

## 5. Conclusion

A model was trained on the AFRIFASHION1600 dataset in this study to rightly classify the gender of Africans by identifying their clothing.

The model achieved a satisfactory accuracy of 87% on the test set given the low amount of training samples.

Much more training and evaluation samples that balances the dataset and represents diverse African cultures would be required to achieve a more relevant and innovative gender classification system.

## Abbreviations

| | |
|---|---|
| RGB | Red, Green, Blue (A Color Model Used for Digital Images) |
| CSV | Comma-Separated Values (A File Format Used for Storing Tabular Data) |
| VGG16 | Visual Geometry Group 16 (A Convolutional Neural Network Architecture) |
| CNN | Convolutional Neural Network (A Type of Deep Learning Model Used for Image Recognition and Classification) |
| Softmax | A mathematical Function Used for Multi-class Classification Problems |
| F1 Score | A statistical Measure of a Model's Accuracy That Considers Both Precision and Recall |
| MNIST | Modified National Institute of Standards and Technology (A Dataset Used for Image Classification) |
| AFRIFASHION1600 | A Dataset of African Traditional Attires Designed for Computer Vision Research |
| OpenCV | Open-Source Computer Vision Library (A Library of Programming Functions for Computer Vision Tasks) |
| Imagenet | A Large-Scale Dataset Used for Image Recognition Challenges |

## Author Contributions

Samuel Ozechi is the sole author. The author read and approved the final manuscript.

## Conflicts of Interest

The author declares no conflicts of interest.